\documentclass[lettersize,journal]{IEEEtran}
\usepackage{amsmath}
\usepackage{amsfonts}
\usepackage{algorithmic}
\usepackage{algorithm}
\usepackage{array}
\usepackage[caption=false,font=normalsize,labelfont=sf,textfont=sf]{subfig}
\usepackage{textcomp}
\usepackage{stfloats}
\usepackage{float}
\usepackage{url}
\usepackage{verbatim}
\usepackage{graphicx}
\usepackage{cite}
\usepackage{amssymb}
\usepackage{multirow}
\usepackage{multicol}
\usepackage{enumitem}
\usepackage{cleveref}
\usepackage{color}
\usepackage{tabularx}
\usepackage{enumitem}
\usepackage{makecell}
\usepackage{tabularx}
\usepackage{amssymb}
\begin{document}

\title{Multiview Manifold Evidential Fusion for PolSAR Image Classification}

\author{Junfei Shi ~\IEEEmembership{Senior Member,IEEE}, Haojia Zhang, Haiyan Jin ~\IEEEmembership{Member,IEEE}, Junhuai Li ~\IEEEmembership{Member,IEEE}, Xiaogang Song ~\IEEEmembership{Member,IEEE}, Yuanfan Guo, Haonan Su, Weisi Lin ~\IEEEmembership{Fellow,IEEE} \vspace{-2em}

\thanks{ Junfei Shi, Haojia Zhang, Haiyan Jin, Junhuai Li, Xiaogang Song, Yuanfan Guo and Haonan Su were the Department of Computer Science and Technology, Shaanxi Key Laboratory for Network Computing and Security Technology, Xi'an University of Technology, Xi'an, China. Corresponding author: Weisi Lin. WSLin@ntu.edu.sg.}
\thanks{Weisi Lin was with the College of Computing and Data Science, Nanyang Technological University, Singapore 639798.}
}


\markboth{Journal of \LaTeX\ Class Files,~Vol.~14, No.~8, August~2021}%
{Shell \MakeLowercase{\textit{et al.}}: A Sample Article Using IEEEtran.cls for IEEE Journals}


\maketitle

\begin{abstract}
Polarimetric Synthetic Aperture Radar (PolSAR) covariance matrices and their extracted multi-features—such as scattering angle, entropy, texture, and boundary descriptors—provide complementary and physically interpretable information for image classification. Traditional fusion strategies typically concatenate these features or employ deep learning networks to combine them. However, the covariance matrices and multi-features, as two complementary views, lie on different manifolds with distinct geometric structures. Existing fusion methods also overlook the varying importance of different views and ignore uncertainty, often leading to unreliable predictions.
To address these issues, we propose a Multiview Manifold Evidential Fusion(MMEFnet) method to effectively fuse these two views. It gives a new framework to integrate PolSAR manifold learning and evidence fusion into a unified architecture. Specifically, covariance matrices are represented on the Hermitian Positive Definite (HPD) manifold, while multi-features are modeled on the Grassmann manifold. Two different kernel metric learning networks are constructed to learn their manifold representations. Subsequently, a trusted multiview evidence fusion, replacing the conventional softmax classifier, estimates belief mass and quantifies the uncertainty of each view from the learned deep features. Finally, a Dempster–Shafer theory-based fusion strategy combines evidence, enabling a more reliable and interpretable classification.
Extensive experiments on three real-world PolSAR datasets demonstrate that the proposed method consistently outperforms existing approaches in accuracy, robustness, and interpretability.

\end{abstract}


\section{Introduction}
Polarimetric Synthetic Aperture Radar (PolSAR) is a vital imaging technology in remote sensing, as it can capture rich information about land surface scattering mechanisms. Unlike optical sensors, PolSAR systems operate in all weather and lighting conditions\cite{r1}\cite{r2}, making them highly reliable for applications such as land cover mapping, urban monitoring, agriculture, and disaster assessment\cite{r3}\cite{r4}\cite{r5}. Accurate classification of PolSAR data allows a detailed understanding of the structure and properties of the terrain, which is essential for environmental monitoring\cite{r6}, resource management and geospatial analysis.


Recently, deep learning has been widely used in the field of PolSAR image since it can learn features automatically. Various architectures, such as Convolutional Neural Networks (CNNs)\cite{r15}\cite{r16}, Recurrent Neural Networks (RNNs)\cite{r17}, and Graph Convolutional Networks (GCNs)\cite{r18}, have been applied to exploit spatial and polarimetric features from SAR data. Existing deep learning-based methods generally convert the complex-valued polarimetric covariance matrix into a real-valued vector form. This transformation enables the direct use of convolutional neural networks (CNNs)\cite{r19} or other standard architectures, which are typically designed for real-valued data, such as 3DCNN, CNN-Transformer, FGCN-CNN\cite{RN41}, PolSAR-KDAC\cite{RN77}, etc. To learn phase information, many complex-valued network structures were proposed by converting the covariance matrix into a complex-valued vector as the network input, such as ST-CNN\cite{RN44}, attention-based CVCNN\cite{r21}, few-shot CNN\cite{10700822}, etc. These methods aim to extract discriminative features from the original polarimetric data without relying on hand-crafted representations. Although effective to some extent, such approaches are inherently limited in scope—they only exploit the second-order scattering statistics encoded in the covariance matrix and often ignore other critical aspects of the scene\cite{r23}, such as texture, structural boundaries or higher-order interactions. 

To address this, some alternative approaches have proposed learning from multiple scattering features, which are derived from polarimetric decomposition techniques (e.g., Cloude-Pottier, Freeman-Durden, Yamaguchi)\cite{r24}\cite{r25}\cite{r26}. These features describe physical scattering mechanisms such as surface, volume, and double-bounce reflections, and provide semantically meaningful cues for classification. For example, Zou et al.\cite{r27} used a combination of alpha-entropy-anisotropy parameters or decomposed power components as input to deep networks. Zhang et al.\cite{10798462} gave a multifrequency fusion method for PolSAR images based on the scattering mechanism. Zhang et al.\cite{10891635} proposed a multilevel conditional diffusion model for PolSAR image classification, treating scattering features as the prior guidance. Although these feature-based models capture additional physical properties beyond the covariance matrix, they typically discard the raw data structure and rely solely on manually engineered descriptors, which may be suboptimal in complex or mixed scenes.

Most existing methods rely on a single-view representation, either using the covariance or coherency matrix directly or focusing solely on derived scattering features. While these two types of features provide complementary information, they also contain some redundant elements. The challenge, therefore, lies in how to effectively fuse these features to extract the most discriminative information.

In the context of deep learning, PolSAR feature fusion methods can generally be divided into two main types. The first approach combines features before they are fed into the deep network\cite{8994163,10569015}. This method concatenates the features along the feature dimension and then performs feature selection automatically. For example, Shi et al.\cite{rs17081422} proposed a multi-feature lightweight DeepLabV3+ network to deal with PolSAR large scene images. Yang et al.\cite{9600819} proposed a reconstruction error-based decomposition method for PolSAR image classification, which selected different subnets to learn their complementary information. However, these approaches treat all features as equally important during fusion, disregarding their varying degrees of relevance. As a result, the model may inadvertently learn less significant or even incorrect features, leading to inaccurate classifications.

The second approach fuses features after feature learning\cite{8771134,10950358}. In this case, features are first learned individually and then combined, either by concatenation or through an attention mechanism, before being passed to a softmax function to output class probabilities. Examples include MCFCN\cite{r28}, CDFNet\cite{10770247} and the attention-based feature selection network\cite{9198918}. However, these methods rely on a softmax classifier, which forces the model to make a hard decision by assigning a high probability to a particular class, even when the predicted class is uncertain or misclassified.
For instance, if the class probabilities are $\{0.4, 0.3, 0.3\}$ across three classes, the softmax would still assign class 1 as the final prediction, although it may be the wrong one. This kind of misprediction is more likely to happen when speckle noises are present in the data.

Therefore, existing fusion methods still face two essential issues.
\begin{itemize}
\item Existing methods convert covariance matrix into a vector, and cascade with other scattering features together, which ignores their different manifold geometric, and result in inaccurate measurement. Specifically, covariance matrices are elements on the Hermitian Positive Definite (HPD) manifold, while multi-features are represented as $m \times n$ matrices on the Grassmann manifold. These two types of data possess fundamentally different geometric structures. Combining them together cannot learn their respective manifold representation well. 
\item Existing methods regard all features trustworthy and classify them with a hard softmax classifier, which may be overly confident in their predictions, regardless of the reliability of the input source. This overconfidence can be particularly problematic when integrating information from multiple uncertain sources, such as in the presence of speckle noise\cite{r29}. So, the result may be unreliable.
\end{itemize}
Consequently, effectively integrating multiple representations within a unified framework remains a significant challenge, particularly when the views lie on distinct manifolds or geometric spaces\cite{r30}. This highlights the need for multiview joint learning strategies that can leverage heterogeneous information to achieve more robust and accurate classification.

Dempster-Shafer (DS) evidence theory has proven to be an effective framework for fusing multiview data\cite{r31}, especially in contexts where uncertainty and conflict between sources must be explicitly modeled\cite{r32}. Unlike traditional methods with hard decision, DS theory defines the belief mass and uncertainty mass for each source, making it more flexible in combining heterogeneous and uncertain data. It can make a trustworthy decision by fusing conflicting evidence in a mathematically reasonable manner. This ability to resolve conflicts and manage uncertainty leads to more robust and accurate classification results, particularly in scenarios where features come from different sources with varying reliability.

Based on this theory, we propose a Multiview Manifold Evidential Fusion Network (MMEFNet) for PolSAR image classification. Here, we consider the original covariance matrix and multi-features as two complementary views. According to their different manifold distributions, two manifold representations are defined: one from covariance matrices in the Hermitian Positive Definite (HPD) manifold, and another from multi-feature matrices in the Grassmann manifold\cite{r33}. Specifically, HPD-based and Grassman-based sGCN (superpixel-based Graph Convolution Network) models are constructed to learn discriminative features, and their outputs are interpreted as evidence vectors that represent support degree for each class. Two sGCN networks can learn manifold-aware geometric separative features from complicated multi-dimension data. The outputted features are evidence vectors, and then transformed into Dirichlet distributions, enabling the estimation of the belief mass and the uncertainty mass\cite{r34}. Then, the DS combination rule is used to fuse the beliefs from both views, adjusting for consistency, and suppressing conflict when the predictions disagree. This fusion is based on mathematical logic and results in a robust class prediction with an associated uncertainty score, enabling more trustworthy decision-making in complex or ambiguous regions. Our approach not only enhances classification performance, but also improves interpretability and reliability, particularly in heterogeneous PolSAR scenes. Therefore, the main contribution of the proposed method can be summarized in three aspects as follows:

1)We propose a trusted MMEFNet for PolSAR image classification, which integrates manifold-based graph learning and DS evidence fusion into a unified framework for the first time. It can effectively combine the advantages of both manifold discriminating ability of GCN and uncertainty modeling ability of DS fusion.

2)The model leverages two complementary views: covariance matrices on the HPD manifold and multi-feature representations on the Grassmann manifold. Each view is modeled by a dedicated sGCN to extract manifold-aware discriminative features, which are then interpreted as Dirichlet-based evidence distributions. 

3)By employing the DS combination rule, the model fuses belief and uncertainty from both views at the evidence level, effectively handling prediction conflicts and improving decision robustness. This approach enhances classification accuracy while providing uncertainty-aware and interpretable outputs, especially beneficial in heterogeneous and ambiguous PolSAR scenes.

The remainder of the paper is organized as follows. Section II presents the background. The proposed method is introduced in Section III in detail. Experimental results and analysis are exhibited in Section IV. Section V is the conclusion.

\section{BACKGROUND}
\subsection{Riemannian Manifold}
\subsubsection{HPD Manifold}

In radar detection and image recognition, Hermitian Positive Definite (HPD) matrices are widely used to model second-order statistics. The Riemannian geometry of the HPD manifold allows for robust distance-based comparisons. A commonly used metric is the Riemannian distance, defined as:
\begin{equation}
D_{\mathrm{RD}}^2(R_1, R_2) = \left\| \log(R_2^{-1/2} R_1 R_2^{-1/2}) \right\|_F^2
\end{equation}
However, due to the computational complexity involved, the Log-Euclidean distance is often preferred:
\begin{equation}
D_{\mathrm{LE}}^2(R_1, R_2) = \left\| \log(R_1) - \log(R_2) \right\|_F^2
\end{equation}
The Log-Euclidean metric enables efficient computation and supports closed-form solutions for averaging on the manifold, making it suitable for detection and classification tasks under uncertainty.

\subsubsection{Grassmann Manifold}

In image set classification, each image set is typically represented as a linear subspace, and the collection of such subspaces forms a Grassmann manifold. A widely used metric is the projection distance, defined as:
\begin{equation}
d(Y_1Y_1^T, Y_2Y_2^T) = \frac{1}{\sqrt{2}} \left\| Y_1Y_1^T - Y_2Y_2^T \right\|_F
\end{equation}
This distance approximates the geodesic between subspaces and can be used to define a projection kernel:
\begin{equation}
k_p(Y_1Y_1^T, Y_2Y_2^T) = \mathrm{tr}(Y_1Y_1^T Y_2Y_2^T) = \|Y_1^T Y_2\|_F^2
\end{equation}
Such kernels are well-defined and widely applied in manifold-based classification.

\subsection{Dempster-Shafer Theory of Evidence (DST)}

The Dempster-Shafer Theory of Evidence (DST), as a significant extension of generalized Bayesian inference, relaxes the strict requirement of precise probability distributions in traditional Bayesian frameworks. It enables uncertainty modeling and reasoning within \emph{interval bounds} defined by upper and lower probabilities. The theoretical foundation of DST originates from the work of \textit{Dempster (1968)}, who introduced \emph{multivalued mappings} and \emph{population space modeling}, and further incorporates the upper and lower probability framework originally proposed by \textit{Boole (1854)}. This gives rise to an inference scheme centered around the notion of \emph{basic belief assignment}.

Dempster-Shafe's evidence fusion treats multiviews as evidences for making a decision. The essence of the fusion process is to synthesize these evidences based on logical rules and ultimately form a comprehensive degree of belief. The Dempster rule of combination is a fundamental mechanism in DST to aggregate two independent sources of evidence defined in a common framework of discernment $\Theta$. The rule is defined as follows:

\begin{equation}
m(A) = \frac{1}{1 - K} \sum_{B \cap C = A} m_1(B) \cdot m_2(C)
\end{equation}

where the conflict factor $K$ is given by:

\begin{equation}
K = \sum_{B \cap C = \emptyset} m_1(B) \cdot m_2(C)
\end{equation}

Here, $m_1$ and $m_2$ represent the belief mass functions from two evidence sources. The value of $K$ reflects the overall \emph{degree of conflict} between the two sources.

\section{Proposed Method}
\begin{figure*}[htbp]
	\centerline{\includegraphics[scale=0.48]{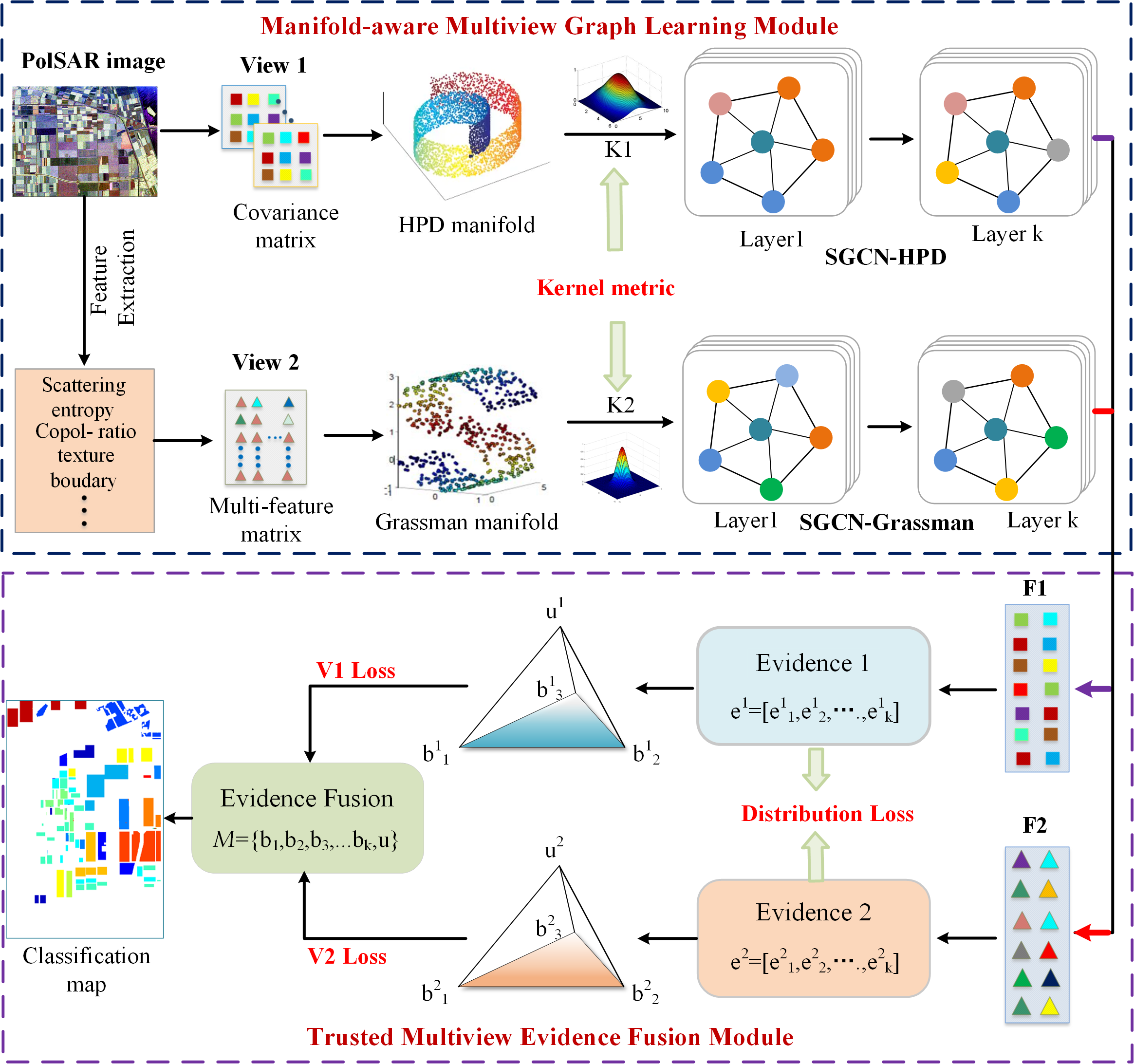}}
	\caption{Framework of the proposed multiview manifold evidential learning network for PolSAR image classification.}
	\label{fig1}
\end{figure*}

In this paper, we propose a multiview manifold evidential fusion network for PolSAR image classification, as illustrated in Fig. \ref{fig1}. The proposed framework treats the original covariance matrices and the derived multi-features as two complementary views. Specifically, two distinct manifold-embedded sGCN models are constructed to extract discriminative features from each view: one operating on the Hermitian Positive Definite (HPD) manifold for covariance matrices, and the other on the Grassmann manifold for multi-feature representations. To construct the graphs, tailored manifold kernel metrics are designed to measure the similarity between nodes in each view, which are then used as edge weights in the respective graphs. The HPD-sGCN and Grassmann-sGCN networks are employed to learn the manifold-aware features from each graph. Finally, an evidence fusion scheme based on Dempster–Shafer theory is designed to effectively integrate complementary information from both views, resulting in improved classification performance with enhanced reliability and interpretability.

\subsection{Multi-view image representation}

Polarimetric Synthetic Aperture Radar (PolSAR) data provide rich scattering information essential for accurate land cover classification. The covariance matrix captures the complete polarimetric scattering characteristics, while manually or automatically extracted multi-features describe additional scattering, statistical, textural, and morphological properties. In order to jointly learn the original data and multi-feature information, we propose a multiview joint graph learning framework that treats the covariance matrix and multi-feature representations as two complementary views of PolSAR data. By integrating these views within a unified learning mechanism, the proposed model enhances discriminative representation of ground objects, thus improving classification performance.

\subsubsection{First View: Covariance Matrix Representation in HPD Manifold}

Unlike conventional single-polarization SAR system, PolSAR transmits and receives horizontally (H) and vertically (V) polarized electromagnetic waves, capturing the full polarimetric information of ground objects. The received signals are represented as a complex scattering matrix:

\begin{equation}
{\bf{S}}= \left[ {\begin{array}{*{20}{c}}
{{S_{hh}}}&{{S_{hv}}}\\
{{S_{vh}}}&{{S_{vv}}}
\end{array}} \right]
\end{equation}
where $S_{hv}$ denotes the complex scattering coefficient with horizontal polarization transmission and vertical polarization reception, a similar concept for $S_{hh}$,$S_{vh}$ and $S_{vv}$. Under the reciprocity assumption ${S_{hv}} = {S_{vh}}$. Then, the scattering matrix can be vectored as $k = \left[ {{S_{hh}},\sqrt 2 {S_{hv}},{S_{vv}}} \right]$. After multilook processing, a covariance matrix can be derived by:
\begin{equation}
{\bf{C}} = k \times {k^H} = \left[ {\begin{array}{*{20}{c}}
{{C_{11}}}&{{C_{12}}}&{{C_{13}}}\\
{{C_{21}}}&{{C_{22}}}&{{C_{23}}}\\
{{C_{31}}}&{{C_{32}}}&{{C_{33}}}
\end{array}} \right]
\end{equation}

This covariance matrix is complex Hermitian positive definite (HPD), which does not reside in a Euclidean space but instead lies in an HPD Riemannian manifold, denoted by $S_{++}^n$. Since this geometric space is nonlinear, Euclidean operations (e.g., averaging, distance computation) may produce invalid or biased results when directly applied. Riemannian geometry metric can be applied on the HPD manifold.


\subsubsection{Second View: Multi-feature representation in Grassman Manifold}

In addition to the polarimetric covariance matrix, we extract a set of complementary features that provide a higher-level description of the land surface scattering behavior. The extracted features are summarized in Table \ref{t1}. These features include:

\begin{table*}[ht]
	\begin{center}
		\caption
		{ \label{t1}{57-dimension features list}}
		\begin{tabular}{|p{2.8cm}|p{3cm}|p{7.50cm}|p{0.90cm}|}
			\hline
			Feature& Feature name& Feature parameter& number\\
			\hline
			\multirow{3}{*}{}& {\makecell[l]{Scattering matrix \\ elements}}& ${real(S_{hh}),img(S_{hh}),real(S_{hv}),img(S_{hv})}$,
			
			${real(S_{vv}),img(S_{vv})}$& 6\\
			\cline{2-4}
			~& {\makecell[l]{Coherency matrix \\ elements}}                                               & ${T_{11},T_{22},T_{33},real(T_{12}),img(T_{12}),real(T_{13})}$,
			
			${img(T_{13}),real(T_{23}),img(T_{23})}$                  & 9      \\
			\cline{2-4}
			~& SPAN                                                                      & $span = {\left| {{S_{hh}}} \right|^2} + {\left| {{S_{hv}}} \right|^2} + {\left| {{S_{vv}}} \right|^2}$                  & 1      \\
			\multirow{5}{*}{\makecell[l]{Target decomposition \\ features}}& {\makecell[l]{Cloud and Pottier \\ decomposition}}& $H$,$A$,$\alpha$                  & 3      \\
			\cline{2-4}
			~& Freeman decomposition                                                     & 	the surface, double-bounce and volume scattering power                  & 3      \\
			\cline{2-4}
			~& Huynen decomposition                                                      & $A_0$,$B_0$,$B,C,D,E,F,G,H$                  & 9      \\
			\cline{2-4}
			~& Co-polarization ratio                                                     & ${r_o} = \frac{{\left\langle {{S_{vv}}S_{vv}^*} \right\rangle }}{{\left\langle {{S_{hh}}S_{hh}^*} \right\rangle }}$                  & 1      \\
			\cline{2-4}
			~& Cross-polarization ratio                                                  & ${r_x} = \frac{{\left\langle {{S_{hv}}S_{hv}^*} \right\rangle }}{{\left\langle {{S_{hh}}S_{hh}^*} \right\rangle }}$                  & 1      \\
			\hline
			\multirow{6}{*}{\makecell[l]{Textural and contour \\ features}} & \multirow{4}{*}{GLCM features}                                            & $Contrast:con = \sum\limits_i {\sum\limits_j {{{(i - j)}^2}} } P\left( {i,j} \right)$                  & 4      \\
			\cline{3-4}
			~&                                                                           & $Energy:Asm = \sum\limits_i {\sum\limits_j {P{{\left( {i,j} \right)}^2}} } $                  & 4      \\
			\cline{3-4}
			~&                                                                           & $Entropy:Ent =  - \sum\limits_i {\sum\limits_j {P\left( {i,j} \right)\log } } P\left( {i,j} \right)$                  & 4      \\
			\cline{3-4}
			~&                                                                           & $Relativity:Corr = {\raise0.7ex\hbox{${\left[ {\sum\limits_i {\sum\limits_j {((i,j)p(i,j)) - {\mu _x}{\mu _y}} } } \right]}$} \!\mathord{\left/
					{\vphantom {{\left[ {\sum\limits_i {\sum\limits_j {((i,j)p(i,j)) - {\mu _x}{\mu _y}} } } \right]} {{\sigma _x}{\sigma _y}}}}\right.\kern-\nulldelimiterspace}
				\!\lower0.7ex\hbox{${{\sigma _x}{\sigma _y}}$}}$                  & 4      \\
			\cline{2-4}
			~& \multirow{2}{*}{\makecell[l]{Edge-line energy \\ features}}                              & ${E_{edge}} = {\raise0.7ex\hbox{${\left( {\frac{1}{n}\sum\limits_{i = 1}^n {{w_i}{x_i}} } \right)}$} \!\mathord{\left/
					{\vphantom {{\left( {\frac{1}{n}\sum\limits_{i = 1}^n {{w_i}{x_i}} } \right)} {\left( {\frac{1}{m}\sum\limits_{j = 1}^m {{w_j}{x_j}} } \right)}}}\right.\kern-\nulldelimiterspace}
				\!\lower0.7ex\hbox{${\left( {\frac{1}{m}\sum\limits_{j = 1}^m {{w_j}{x_j}} } \right)}$}}$                  & 4      \\
			\cline{3-4}
			~&                                                                           & ${E_{line}} = \min \{ E_{edge}^{ij},E_{edge}^{jk}\}$                  & 4      \\
			\hline
			Total                                          &                                                                           &                   & 57 \\
			\hline
		\end{tabular}
	\end{center}
\end{table*}

(a) Scattering features: These features are derived from the original data or physical decomposition methods, including the vectorization of the original scattering matrices, as well as cloude-Pottier\cite{10335659}, Freeman-Durden\cite{r8}, and Huynen decompositions\cite{RN74}, etc., as illustrated in Table \ref{t1}. These features capture the dominant scattering mechanisms: surface, double-bounce, and volume, thus enhancing semantic interpretability.

(b) Textural features: Gray-level co-occurrence matrix (GLCM)\cite{RN75} is used to calculate statistical descriptors over local neighborhoods, these features characterize spatial patterns and roughness, which are crucial for distinguishing between classes with similar scattering signatures.

(c) Boundary and shape features\cite{RN76}: They include edge strength, boundary contrast, and object morphology metrics, which are useful to delineate object contours and spatial structures, especially in heterogeneous areas.

To integrate these heterogeneous descriptors into a unified mathematical form, we represent them as a feature vector. Thus, for each superpixel $S_i$, its mean feature vector is computed as: 
\[
\bar{x}_i = \frac{1}{n_i} \sum_{j=1}^{n_i} x_{ij}
\]
where $x_{ij}$ denotes the feature vector of the $j$-th pixel within the superpixel $S_i$, and $n_i$ is the number of pixels in $S_i$. To learn the correlation of multiple features, the covariance matrix is defined by:
\[
C_i = \frac{1}{n_i} \sum_{j=1}^{n_i} (x_{ij} - \bar{x}_i)(x_{ij} - \bar{x}_i)^H
\]

This matrix characterizes the second-order spectral statistics within the superpixel. Based on this, we perform the eigenvalue decomposition (or singular value decomposition) on each covariance matrix as follows:
\[
C = U \Lambda U^H
\]
where $U$ is an orthogonal matrix composed of eigenvectors. The eigenvectors corresponding to the top $q$ largest eigenvalues are selected to form an orthonormal basis matrix $U_q$, which satisfies $U_q^H U_q = I_q$
The $q$-dimensional subspace spanned by the selected orthonormal basis defines a point on the Grassmann manifold, representing each superpixel. This yields $N$ Grassmannian points—each stored as a $d\times q$ orthonormal matrix—forming the second view. This manifold representation enables robust subspace-based comparison and learning, inherently capturing geometric invariance and reducing sensitivity to noise or scaling variations across features.

\subsection{Multiview SGCN Model with Manifold Embedding}

In this section, we propose a novel multiview sGCN framework designed to effectively learn from two structurally distinct representations of PolSAR data: the covariance matrix (View 1) and the multi-feature matrix (View 2). Two views reside in different manifold geometries and provide complementary discriminative information. The core idea is to construct two separate yet coordinated sGCNs that operate on graphs built from these two views, where manifold-aware similarity metrics guide both graph construction and feature learning.

To reduce noise and preserve spatial coherence, the image is first oversegmented into superpixels by the Pol\_ASLIC method\cite{7858788}, which can obtain regular superpixels with similar size and reduce speckle with polarimetric features. Each superpixel is treated as a graph node, and adjacent nodes have an edge. The graph is defined as: $G =\{V,E,A\}$, where$V$ is the node set, $E$ is the edge set, and $A$ is the adjacent matrix.

\subsubsection*{Manifold Kernel Metric Learning for Two Views}

To consider the adjacent relationship, we define a weighted adjacent matrix $A$, where the similarity of two adjacent nodes is calculated as the weight. Since two views are endowed with different manifold spaces, different manifold kernel metrics are defined for two views.

For the first view, each node in the graph corresponds to a superpixel, represented by a set of Hermitian Positive Definite (HPD) covariance matrices. To properly measure the similarity between two graph nodes $i$ and $j$, we define a manifold kernel metric based on the log-Euclidean Riemannian distance:
\begin{equation}
d_{\text{HPD}}(C_i, C_j) =\left\| \log(C_i) - \log(C_j) \right\|_F
\end{equation}
This distance is then used to construct the adjacency matrix $\mathbf{A}^{(1)}$ of the graph corresponding to the covariance matrix view, where weights reflect pairwise similarities on the HPD manifold:
\begin{equation}
A_{ij}^{(1)} = \exp\left( -\frac{d_{\text{HPD}}^2(C_i, C_j)}{\sigma^2} \right)
\end{equation}

For the second view, each node is represented by a multi-feature matrix, composed of scattering, textural, and boundary features. Feature matrices are projected onto a Grassmann manifold $\mathcal{G}(n,q)$, where the matrix dimension is $n\times q$. Each point represents a subspace.

The similarity between nodes is computed using a Grassmann kernel, such as the projection kernel:
\begin{equation}
A_{ij}^{(2)} =k_{\text{Grass}}(U_i, U_j) = \left\| U_i^\top U_j \right\|_F^2,
\end{equation}
where $\|\cdot\|_F$ denotes the Frobenius norm. This kernel reflects the degree of overlap and structural similarity between two subspaces on the Grassmann manifold. 

This kernel is used to define the adjacency matrix $\mathbf{A}^{(2)}$ for the graph of View 2. Finally, the two $N \times N$ kernel matrices serve as geometry-aware similarity measures, which are further integrated into the graph structure modeling and feature propagation of the multiview sGCN.
\subsubsection*{Manifold-aware Multi-view sGCN Model}

Each view independently constructs an undirected, weighted graph, where edges are established based on a manifold-aware similarity function that captures the intrinsic geometric relationships between nodes. Specifically, for two views, different manifold kernel metrics are calculated as the similarities of two nodes. These similarity values are encoded in the graph adjacency matrix $\mathbf{A}$, where higher weights indicate stronger connections between two nodes in semantic or scattering similarity. As a result, the sGCN can propagate information through the graph in a geometry-consistent manner, improving the network’s ability to model local and global contextual dependencies. In addition, a mutual information loss is defined to ensure the consistency of class labels across views.

Based on the above, kernel matrices are constructed respectively for the HPD and Grassmann manifolds. Two parallel sGCN branches are constructed. For each sGCN, the weighted adjacency matrix $A^{v}$ is calculated using the manifold kernel metric and $v$ is the number of views. Then, multi-layer graph convolution operations are then performed on these structure-aware graph:
\[
H^{(l+1, v)} = \sigma\left( \hat{D}^{-1/2} A^{(v)} \hat{D}^{-1/2} H^{(l, v)} W^{(l)} \right)
\]
\[H^{(0, v)} = X^{(v)}\]
where $X^{(v)}$ is the input feature for view $v$. $W^l$ is the weight matrix in $l$th hidden layer. Finally, the output feature is mapped from superpixel back to the pixel space to obtain pixelwise graph features:
\[
F_{\text{GCN}}^{(v)} = Q H^{(L, v)} \in \mathbb{R}^{HW \times d}
\]
where Q is the projection matrix, in which $Q_{ij}=1$ when the pixel $i$ is within the superipxel $j$, otherwise 0. Thus, features for two views can be extracted from the graph convolution branches, expressed as:
\[
 \left[ \textbf{F}_{\text{view1}} \, ; \, \textbf{F}_{\text{view2}} \right]
\]
This process enables multi-view feature learning enhanced by manifold-aware graph structures, providing a unified and structured input representation for trusted classification. Each view independently models the HPD matrix or multi-feature relationships within its respective manifold space, while maintaining consistent network architecture to ensure alignment and robustness in feature fusion.

\subsection{Multiview Evidential Fusion Module for Robust Classification}

Both the HPD-sGCN and Grassmann-sGCN branches capture complementary discriminative features, while they may sometimes yield inconsistent or uncertain predictions due to noisy measurements (e.g., speckle in PolSAR), occlusions or low-texture areas, and intra-class variability across views. A simple softmax average cannot handle uncertainty or contradiction between views. Thus, a trusted fusion strategy is required. To achieve robust and trustworthy classification, we introduce an evidence-based fusion mechanism grounded in subjective logic and Dempster–Shafer theory. This allows the system to handle uncertainty and conflicts between views in a principled way. It can offer more precise uncertainty quantification and enables flexible integration of multiple views to support reliable decision-making.

In the aforementioned dual-branch sGCN architecture, we extract manifold-aware feature representations $\mathbf{F}_{view1}$ and $\mathbf{F}_{view2}$ from two views, respectively. To improve the reliability and robustness of the model prediction, we introduce a multi-view fusion strategy based on Dempster-Shafer Theory (DST), which integrates the classification evidence from each view at the evidence level, thereby constructing a trustworthy multi-view classification model. Notably, \textit{evidence} denotes the information extracted from the input that contributes to the classification decision. This evidence serves as the basis for estimating the concentration parameters of a Dirichlet distribution. The Dempster–Shafer theory of evidence assigns belief masses to subsets within a frame of discernment, representing all possible class labels in multi-class classification. It allows uncertainty modeling by distributing beliefs across classes. When multiple sources provide evidence, Dempster’s rule combines shared beliefs to produce a merged belief mass and updated uncertainty.

Specifically, for each pixel location $i$, we first predict its classification evidence vectors $\mathbf{e}_i^{(1)} \in \mathbb{R}^C$ and $\mathbf{e}_i^{(2)} \in \mathbb{R}^C$ by ReLu function  from the network outputs of the two views. Assuming $\textbf{o} = [{o_1},{o_2}, \cdots ,{o_C}]$
is the output of the neural network, $C$ corresponds to the number of classes. Evidence assignment can be defined as
\begin{equation}
{e_i} = ReLu(\textbf{o}) = \left\{ {\begin{array}{*{20}{c}}
   {\begin{array}{*{20}{c}}
   {0} & {{\rm{for }}{{\rm{o}}_i} \le 0}  \\
\end{array}}  \\
   {\begin{array}{*{20}{c}}
   {{o_i}} & {{\rm{for }}{{\rm{o}}_i} > 0}  \\
\end{array}}  \\
\end{array}} \right.
\end{equation}

According to Subjective Logic theory, each evidence vector can be further converted into the corresponding parameters of a Dirichlet distribution as:
\begin{equation}
\boldsymbol{\alpha}_i^{(v)} = \mathbf{e}_i^{(v)} + 1, \quad v \in \{1,2\}
\end{equation}
Subsequently, the belief mass and uncertainty mass for each view can be derived by:
\begin{equation}
b_{i,k}^{(v)} = \frac{e_{i,k}^{(v)}}{S_i^{(v)}}, \quad
u_i^{(v)} = \frac{C}{S_i^{(v)}}, \quad
S_i^{(v)} = \sum_{k=1}^{C} \alpha_{i,k}^{(v)}
\end{equation}

To effectively fuse the decision beliefs from the two views, we adopt Dempster’s combination rule for evidence-level fusion. For the belief sets 
$\mathcal{M}_1 = \{ b_{i,k}^{(1)}, u_i^{(1)} \}$ and 
$\mathcal{M}_2 = \{ b_{i,k}^{(2)}, u_i^{(2)} \}$ derived from the two views at pixel $i$, the combined belief mass and uncertainty mass are computed as follows:
\begin{equation}
\begin{array}{l}
 {b_{i,k}} = \frac{1}{{1 - {C_i}}}\left( {b_{i,k}^{(1)}b_{i,k}^{(2)} + b_{i,k}^{(1)}u_i^{(2)} + b_{i,k}^{(2)}u_i^{(1)}} \right), \quad  \\ 
{u_i} = \frac{{u_i^{(1)}u_i^{(2)}}}{{1 - {C_i}}} \\
 \end{array}
\end{equation}
where the conflict factor is:
\begin{equation}
C_i = \sum_{j \neq k} b_{i,j}^{(1)} b_{i,k}^{(2)}
\end{equation}
measures the inconsistency between the predictions of the two views at the given pixel.

Next, according to the fused belief values $\{b_{i,k}, u_i\}$, we recover the combined evidence as follows:
\begin{equation}
S_i = \frac{C}{u_i}, \quad
e_{i,k} = b_{i,k} \cdot S_i, \quad
\alpha_{i,k} = e_{i,k} + 1
\end{equation}

Finally, the fused Dirichlet distribution $D(\mathbf{p}_i \mid \boldsymbol{\alpha}_i)$ provides both the class probability distribution and uncertainty estimation for each pixel, where the mean of the distribution is taken as the final trustworthy classification output:
\begin{equation}
\hat{y}_{i,k} = \frac{\alpha_{i,k}}{\sum_{j=1}^{C} \alpha_{i,j}}
\end{equation}

Compared to the traditional softmax output, the Dirichlet distribution not only models the class-wise probability distribution but also introduces an overall uncertainty mass, effectively alleviating the issue of overconfidence. In addition, this approach enables the model to dynamically adjust the reliability of different views, thereby enhancing its responsiveness to outlier samples and out-of-distribution data. The multi-view evidence fusion not only improves classification performance but also provides more interpretable decision-making for downstream tasks.

\subsection{Loss Function}
To jointly optimize the proposed multi-view sGCN and the trusted classification module, we design the total loss function as follows:
\begin{equation}
\mathcal{L}_{\text{Total}} = \mathcal{L}_{\text{sGCN}} + \mathcal{L}_{\text{Distribution}}
\end{equation}
where the loss of sGCN adopts the standard cross-entropy form:
\begin{equation}
\mathcal{L}_{\text{sGCN}} = -\sum_{i=1}^{N} \sum_{c=1}^{C} Y_c(i) \log(L_c(i))
\end{equation}
which supervises the classification learning under two manifold views.

The trusted multiview classification module adopts a Dirichlet-based evidential loss function~\cite{r31}, including the cross-entropy terms between the fused Dirichlet distribution $\boldsymbol{\alpha}_i$ and each view-specific distribution $\boldsymbol{\alpha}_i^{(v)}$, as well as a KL divergence regularization with a prior $\tilde{\boldsymbol{\alpha}}_i$. 
For each sample $i$, the loss is defined as:
\begin{equation}
\mathcal{L}(\alpha_i)
=\mathcal{L}_{CE}(\alpha_i)
+\lambda_t\,\mathrm{KL}\!\left[\mathrm{Dir}(\tilde{\alpha}_i)\,\|\,\mathrm{Dir}(\mathbf{1})\right],
\label{eq:dirichlet_loss}
\end{equation}
where the expected cross-entropy term is formulated as:
\begin{equation}
\mathcal{L}_{CE}(\alpha_i)
=\sum_{k=1}^{K}y_{ik}\,[\psi(S_i)-\psi(\alpha_{ik})],
\label{eq:dirichlet_ce}
\end{equation}
with $S_i=\sum_{k=1}^{K}\alpha_{ik}$, and $\psi(\cdot)$ denoting the digamma function.
Here, $\tilde{\alpha}_i=y_i+(1-y_i)\odot\alpha_i$ replaces the Dirichlet parameters of the ground-truth class with 1 to prevent penalizing its evidence.

The KL divergence term between two Dirichlet distributions is computed as:
\begin{equation}
\small
\begin{aligned}
\mathrm{KL}\!\left[\mathrm{Dir}(\tilde{\alpha}_i)\,\|\,\mathrm{Dir}(\mathbf{1})\right]
&=\log\frac{\Gamma(\sum_{k}\tilde{\alpha}_{ik})}{\Gamma(K)}
-\sum_{k=1}^{K}\log\Gamma(\tilde{\alpha}_{ik}) \\
&\quad+\sum_{k=1}^{K}(\tilde{\alpha}_{ik}-1)
\left[\psi(\tilde{\alpha}_{ik})-\psi\!\left(\sum_{j}\tilde{\alpha}_{ij}\right)\right].
\end{aligned}
\label{eq:dirichlet_kl}
\end{equation}

Thus, the overall distribution loss for all samples and views can be written as:
\begin{equation}
\mathcal{L}_{\text{Distribution}}
=\sum_{i=1}^{N}\Big[\mathcal{L}(\alpha_i)
+\sum_{v=1}^{V}\mathcal{L}(\alpha_i^{(v)})\Big].
\label{eq:dirichlet_distribution}
\end{equation}

This objective enables the model to jointly learn discriminative evidence from both fused and view-specific Dirichlet distributions, 
while regularizing non-ground-truth evidence through KL divergence. 
As a result, it improves classification reliability and uncertainty modeling for PolSAR data.

\section{Experimental results and analysis}
\subsection{Experimental Data and Settings}

\begin{figure}
	\centerline{\includegraphics[scale=0.45]{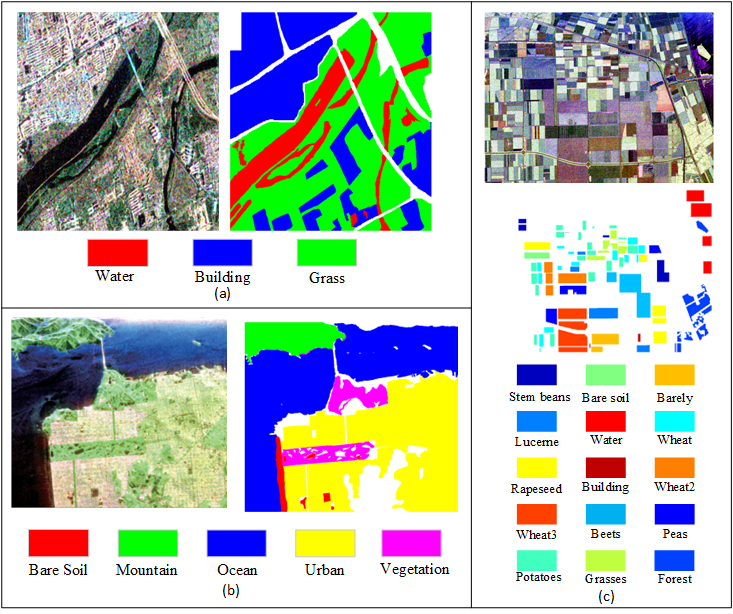}}
	\caption{Dataset. (a)PauliRGB image and label map of Xi'an data set; (b)PauliRGB image and the ground-truth map of San Francisco data set; (c)PauliRGB image and label map of Flevoland data set.}
	\label{fig2}
\end{figure}

To validate the effectiveness of the proposed MMEFnet method, we conducted extensive experiments on three representative PolSAR datasets acquired by different sensors and covering various frequency bands. The key parameters of each dataset are summarized in Table~\ref{t2}, and detailed descriptions are given as follows:

\begin{itemize}
   \item \textbf{\textit{Xi’an Dataset:}} This dataset was acquired by RADARSAT-2 system and consists of C-band fully polarimetric SAR images. The image size is $512 \times 512$ pixels with a spatial resolution of $8 \times 8$ meters. It includes three main land cover classes:water , grassland, and buildings. Fig.~\ref{fig2}(a) shows the corresponding Pauli RGB pseudo-color image and ground-truth label map.

    \item \textbf{\textit{San Francisco Dataset:}} This dataset was collected by the AIRSAR system developed by NASA JPL in the United States. It provides L-band fully polarimetric SAR images with a size of $900 \times 1024$ pixels and a spatial resolution of $8 \times 8$ meters. The image covers five typical land cover types: bare soil, urban area, ocean, mountain, and vegetation. Fig.~\ref{fig2}(b) presents the Pauli RGB image along with its corresponding ground-truth map.
    
    \item \textbf{\textit{Flevoland dataset:}} The last dataset is the Flevoland dataset, acquired on August 16, 1989, by the AIRSAR airborne sensor over the Flevoland region. It is a quad-polarimetric L-band SAR image with four look angles. The image size is 750 × 1024 pixels and includes 15 main types of land cover: bare soil, barely, beets, buildings, forest, grass, luceme, peas, potatoes, rapeseed, stem beans, water, and three different types of wheat. Fig.~\ref{fig2}(c) illustrates its Pauli RGB image and ground truth map.
\end{itemize}
\begin{table}[htbp]
\centering
\caption{Descriptions on Three PolSAR Data Sets}
\label{t2}
\begin{tabular}{lcccc}
\hline
\textbf{Name} & \textbf{System} & \textbf{Band} & \textbf{Dimensions} & \textbf{Class} \\
\hline
Xi’an & RADARSAT-2 & C & 512$\times$512 & 3 \\
San Francisco & AIRSAR & L & 900$\times$1024 & 5 \\
Flevoland & AIRSAR & L & 750$\times$1024 & 15 \\
\hline
\end{tabular}
\end{table}

The model training process was configured with the following key parameters: an initial learning rate of 0.0001, 300 training epochs, a batch size of 64, and the Adam optimizer. For data partition, 5\% of labeled samples from each dataset were randomly selected for training, 1\% were used for validation, and the remaining 94\% were used for testing to ensure fairness and representativeness.  All experiments were conducted on a Windows 10 operating system using the PyTorch 2.5.1 deep learning framework, running on a machine equipped with an Intel Core i7-12700K processor, 32~GB of RAM, and an NVIDIA GeForce RTX 3060 GPU with 12~GB of VRAM.

The effectiveness of the proposed MMEFnet model for PolSAR image classification is evaluated by comparing it with several representative baselines and structural variants, including Super\_RF\cite{6247500} CVCNN\cite{8039431}, PolMPCNN\cite{9424197}, DFGCN\cite{9274334}, SpectralDiff\cite{10234379} and HybridCVNet\cite{10693615}. These methods cover convolutional feature extraction, polarimetric information fusion, graph structure modeling, contextual learning, and multi-modal fusion. Super\_RF employs various polarimetric features to train a random forest and refines results through superpixel-based spatial correction. CVCNN uses a standard CNN to assess the capacity of basic convolutional structures. PolMPCNN introduces multi-scale sampling and polarization orientation modeling to enhance polarimetric feature representation. DFGCN leverages graph convolution and a dynamic feature guidance mechanism to capture spatial context. SpectralDiff combines diffusion processes and attention mechanisms for spectral–spatial modeling, yielding strong classification performance. HybridCVNet integrates CV-CNN and CV-ViT to utilize complex-valued scattering features and complementary information. Performance was quantitatively assessed using five commonly adopted classification metrics: overall accuracy (OA), average accuracy (AA), Kappa coefficient.

\subsection{Experimental Result and Analysis}
Based on the proposed MMEFnet model, three real-world PolSAR datasets acquired by different sensors and spanning diverse land cover types were selected for experimental validation and comparative analysis.

\subsubsection{Xi'an data set}
\begin{figure*}
	\centerline{\includegraphics[scale=0.6]{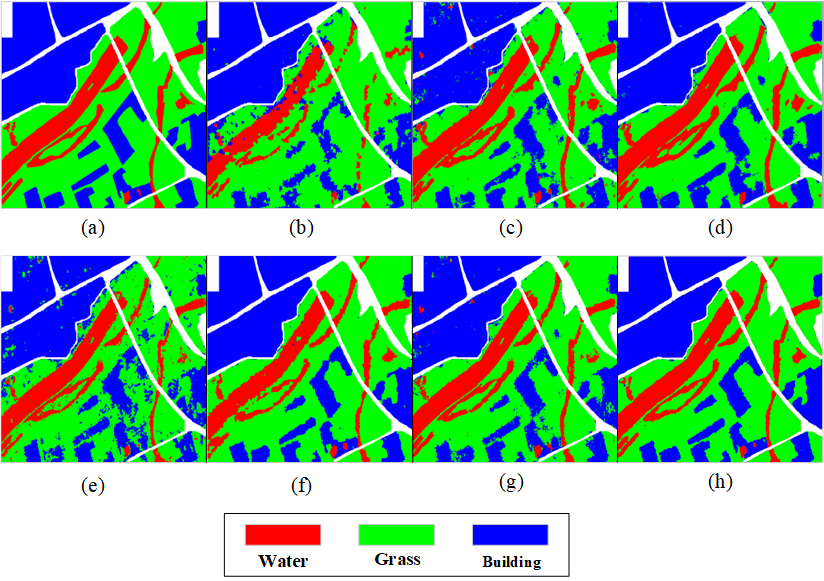}}
	\caption{Classification maps of the Xi'an data set. (a) Ground truth; (b) Super\_RF; (c) CV-CNN; (d) PolMPCNN; (e) DFGCN; (f) SpectralDiff; (g) HybridCVNet; (h) proposed.}
	\label{fig3}
\end{figure*}

As shown in Figs.~\ref{fig3}(b)-(h), different methods exhibit significant discrepancies in land cover classification on the Xi’an dataset. In Fig.~\ref{fig3}(b), the Super\_RF method fails to classify \textit{water} area. CV-CNN demonstrates considerable misclassifications in \textit{building} region, with boundaries often confused with \textit{water} or \textit{grass}, reflecting its limitations in boundary modeling. PolMPCNN shows relatively balanced performance in both homogeneous and heterogeneous areas, but confusion still arises at the edges between \textit{building} and \textit{grass} classes. DFGCN struggles with global context modeling, resulting in scattered predictions in heterogeneous regions and reduced spatial coherence. SpectralDiff effectively suppresses noise in homogeneous regions but performs poorly in boundary areas. HybridCVNet can accurately preserve \textit{water} boundaries but still makes some errors in classifying \textit{grass} and \textbf{buildings}. In contrast, the proposed MMEFnet integrates Riemannian geometry and uncertainty modeling, showing enhanced discrimination at heterogeneous edges and class boundaries. The classification maps are more coherent and accurate, demonstrating strong adaptability and robustness in complex scenarios.

Quantitative results on the Xi’an dataset are reported in Table~\ref{t3}, where the proposed method achieves the best performance across all metrics, including overall accuracy (OA), average accuracy (AA), Kappa coefficient. Compared with Super\_RF, CV-CNN, PolMPCNN, DFGCN, SpectralDiff, and HybridCVNet, MMEFnet improves OA by $7.79\%$, $5.36\%$, $3.72\%$, $8.19\%$, $1.34\%$, and $3.17\%$, respectively. Notably, Super\_RF performs worst in identifying the \textit{water} class, though it shows effective separation of \textit{grass} and \textit{building}, resulting in improved OA. CV-CNN underperforms in the \textit{grass} class with an accuracy of only $90.68\%$, as many \textit{grass} pixels are misclassified as \textit{water}, revealing its limitations in boundary discrimination. DFGCN fails to achieve satisfactory classification in all three categories, especially for \textit{water} ($86.15\%$), due to its insufficient modeling of spatial boundaries. In contrast, PolMPCNN utilizes multi-scale convolution to integrate local and global features, achieving $97.68\%$ accuracy for the \textit{building} class. Furthermore, the HybridCVNet method exhibits balanced performance across various metrics; however, its overall accuracy (OA) remains lower than that of the proposed MMEFnet method. These results clearly demonstrate the effectiveness and advantages of the proposed method in distinguishing different land cover types and enhancing the classification accuracy of PolSAR images.In the tables, the underlined values denote the second-best results, while the bold values indicate the best results.

\begin{table*}[htbp]
\caption{Classification accuracy (\%) of each method on the Xi’an dataset.}
\label{t3}
\centering
\setlength{\tabcolsep}{1.2mm}  
\small  
\begin{tabular}{c|c|c|c|c|c|c|c}
\hline
Class & Super\_RF & CV-CNN & PolMPCNN & DFGCN & SpectralDiff & HybridCVNet & Proposed \\
\hline
Water    & 70.91 & \underline{94.55} & 95.52 & 86.15 & 90.20 & 92.01 & \textbf{94.60} \\
Grass    & 94.97 & 90.68 & 90.95 & 88.36 & \underline{97.23} & 94.99 & \textbf{98.31} \\
Building & 90.94 & 93.81 & 97.68 & 92.65 & \underline{97.84} & 95.05 & \textbf{98.45} \\
\hline
OA       & 89.94 & 92.37 & 94.01 & 89.54 & \underline{96.39} & 94.56 & \textbf{97.73} \\
AA       & 85.61 & 93.01 & 94.71 & 89.05 & \underline{95.09} & 94.02 & \textbf{97.08} \\
Kappa    & 83.02 & 87.51 & 90.25 & 82.75 & \underline{94.02} & 91.02 & \textbf{96.21} \\
\hline
\end{tabular}
\vspace{0.5em}
\end{table*}

\subsubsection{San Francisco data set}
\begin{figure*}
	\centerline{\includegraphics[scale=0.6]{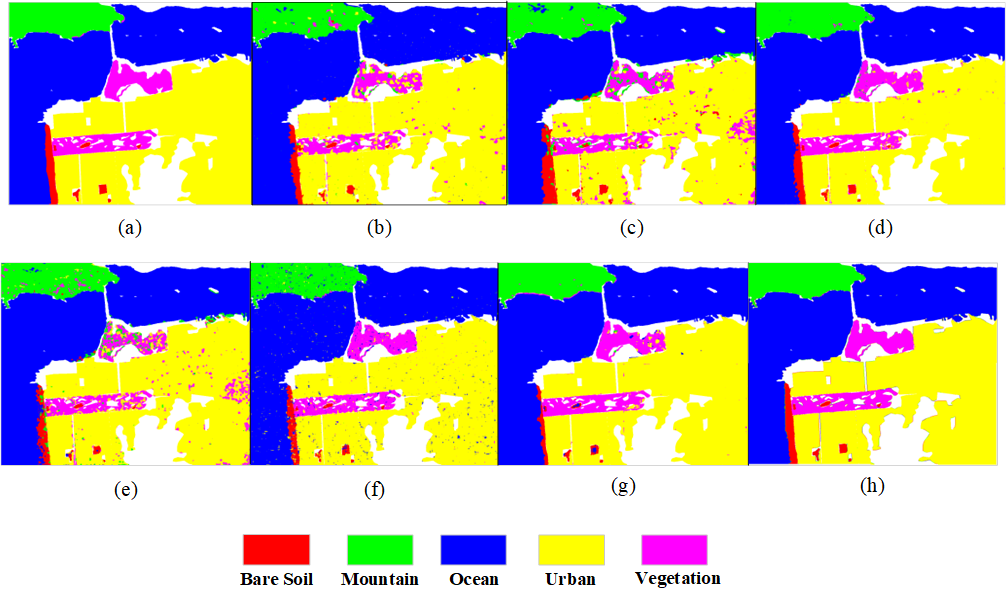}}
	\caption{Classification maps of the San Francisco data set. (a) Ground truth; (b) Super\_RF; (c) CV-CNN;(d) PolMPCNN; (e) DFGCN; (f) SpectralDiff; (g) HybridCVNet; (h) proposed.}
	\label{fig4}
\end{figure*}

Figures ~\ref{fig4}(b)-(h) illustrate the classification results of six comparison methods and the proposed method on the \textit{San Francisco} dataset. It can be observed that Super\_RF exhibits poor classification performance in the \textit{bare soil} and \textit{vegetation} categories, primarily due to its limited ability to extract low-level features that can effectively represent the complex semantic information of these land cover types. CV-CNN and DFGCN show significant pixel-level confusion between the \textit{urban} and \textit{vegetation} classes, indicating evident inter-class ambiguity. PolMPCNN achieves notable performance improvements in all categories; however, local misclassifications still occur between \textit{mountain} and \textit{bare soil} classes. HybridCVNet and SpectralDiff enhance feature representation by incorporating deeper network architectures, resulting in improved accuracy in \textit{vegetation} classification. Nevertheless, HybridCVNet is still under-performed in identifying \textit{bare soil} regions, leading to partial omissions. In contrast, MMEFnet produces classification maps with the clearest boundaries and the strongest spatial consistency among the five land cover categories, demonstrating superior land cover discrimination and semantic representation capabilities.

As shown in Table~\ref{t4}, different methods exhibit varying performance in land cover categories and overall metrics. Super\_RF performs reasonably well in the \textit{mountain} and \textit{urban} classes, but its low accuracy in \textit{vegetation} and \textit{bare soil} limits its overall effectiveness. CV-CNN achieves a relatively high accuracy of 98.51\% in \textit{mountain} class, yet its AA and Kappa are comparatively lower, indicating limitations in boundary discrimination. PolMPCNN performs excellently in the \textit{urban} and \textit{vegetation} classes, with accuracies of 99.73\% and 97.24\%, respectively, achieving the highest OA 97.93\% and Kappa 98.64\%, demonstrating strong overall classification capability. DFGCN shows sub-par performance across all categories, particularly in \textit{vegetation} 68.00\% and \textit{bare soil} 65.62\%, reflecting insufficient modeling of spatial features. SpectralDiff achieves competitive results in the \textit{ocean} and \textit{urban} classes, but its performance in \textit{bare soil} 76.06\% is relatively weak, affecting its overall consistency. HybridCVNet yields good accuracy in several categories, such as \textit{urban} and \textit{ocean}, and achieves an OA of 98.31\%, yet its performance in the \textit{bare soil} class is notably low 52.19\%. In contrast, the proposed MMEFnet achieves the best or near-best accuracy in multiple key classes, including Mountain 99.85\%, \textit{bare soil} 95.82\%, and \textit{vegetation} 98.60\%. It also attains competitive OA 99.34\% and the highest AA 98.44\%, highlighting its ability in robustness and generalization for PolSAR image classification.

\begin{table*}[htbp]
	\caption{\label{t4}Classification Accuracy of Each Comparison Algorithm on the San Francisco Dataset($\%$)}
	\begin{center}
    \setlength{\tabcolsep}{1.5mm}{
		\begin{tabular}{c|c|c|c|c|c|c|c}
			\hline
			class & Super\_RF& CV-CNN& PolMPCNN& DFGCN& SpectralDiff& HybirdCVnet& proposed \\
			\hline
			Bare soil & 77.00 & \underline{83.06} & 73.21 &  65.62 &  76.06 &  52.19 & \textbf{95.82} \\
			 Mountain & 93.98 & 98.51& 94.42 & 92.68 & 98.05 & \underline{99.30} & \textbf{99.85} \\
			Ocean & 99.01 & 96.13 &\underline{99.53}  & 98.01 & 99.09 &\textbf{99.58}  & 98.84 \\
			Urban & 98.36 & 94.25 & \textbf{99.73} & 95.82 & 98.45&\underline{99.67}  & 99.51 \\
            Vegetation &69.83 &76.87 &\underline{97.24} &68.00 & 90.60 &91.24 &\textbf{98.60}\\
			OA & 96.02 & 93.81 & 97.93 & 94.10 & 97.49 & \underline{98.31} & \textbf{99.34} \\
			AA & 87.64 & 89.27 & \underline{93.64} & 84.03 & 91.72 & 88.62 & \textbf{98.44} \\
			Kappa & 93.71& 90.45 & \textbf{98.64} & 94.09 & 96.04 & 97.33 & \underline{98.27} \\
			\hline
		\end{tabular}}
	\end{center}
\end{table*}

\begin{table*}[htbp]
	\caption{\label{t5}Classification Accuracy of Each Comparison Algorithm on the Flevoland Dataset($\%$)}
	\begin{center}
    \setlength{\tabcolsep}{1.5mm}{
		\begin{tabular}{c|c|c|c|c|c|c|c}
			\hline
			class & Super\_RF& CV-CNN& PolMPCNN& DFGCN& SpectralDiff&HybridCVNet& proposed \\
			\hline
			Stem beans & 91.55 & \underline{99.72} & \textbf{99.78} & 99.69 & 99.69 &  99.52 &99.59 \\
			Peas & 81.90& 85.08 & \underline{99.98} & 99.51 & \textbf{100.00} & 99.26 & 99.95 \\
			Forest & 78.64 & 99.81 & \textbf{99.96} &  97.65 & \underline{99.95} & 99.89 & 99.80\\
			 Lucerne & 88.01 & 98.17 & 98.44 & \underline{99.54} & 96.20 & \textbf{99.92} & \underline{99.72} \\
           Wheat & 79.14& 99.22 & 98.01 &98.06 &  \textbf{99.94} & 99.51 & \underline{99.93} \\
           Beets &82.42&98.65&95.51&98.22& 99.20 & \textbf{100.00} &\underline{99.92}\\
        Potatoes&79.77&99.13&98.76&97.71&\textbf{99.94} &97.10&\underline{99.64}\\
        Bare soil& 74.33&\underline{99.93}&95.70&97.53&\textbf{100.00} &97.86&\textbf{100.00}\\
        Grasses&62.00&\textbf{100.00}&99.83&91.51&98.42 &99.84&\underline{99.98}\\
        Rapeseed&53.65&94.18&93.55& 94.24& \textbf{100.00} &88.54&\underline{99.50}\\
        Barely&85.29&99.41&95.96&99.20& \underline{99.75}&92.76&\textbf{100.00}\\
        Wheat2&55.01&\underline{99.58}&40.24&91.45&\textbf{100.00}&99.28&\textbf{100.00}\\
        Wheat3&84.36&99.37&93.88&99.51&\underline{99.92}&99.19&\textbf{99.97}\\
        Water& 97.97&\textbf{100.00}&96.01&99.67&\underline{99.97}& 99.84&\textbf{100.00}\\
        Buildings&86.34&\textbf{99.98}&91.27&80.88 & 82.56 & 91.81&\underline{92.61}\\
			OA &  78.70 & 98.96 & 93.82 & 97.54 & \underline{99.73} & 98.98 & \textbf{99.75} \\
			AA &  78.69 & \underline{99.14} & 93.06 &  96.29 & 98.57&  98.35 & \textbf{99.40} \\
			Kappa & 76.73 & 98.86 & 93.26 & 97.32 & \underline{99.71} & \underline{98.88} & \textbf{99.73} \\
			\hline
		\end{tabular}}
	\end{center}
\end{table*}

\begin{figure*}
	\centerline{\includegraphics[scale=0.6]{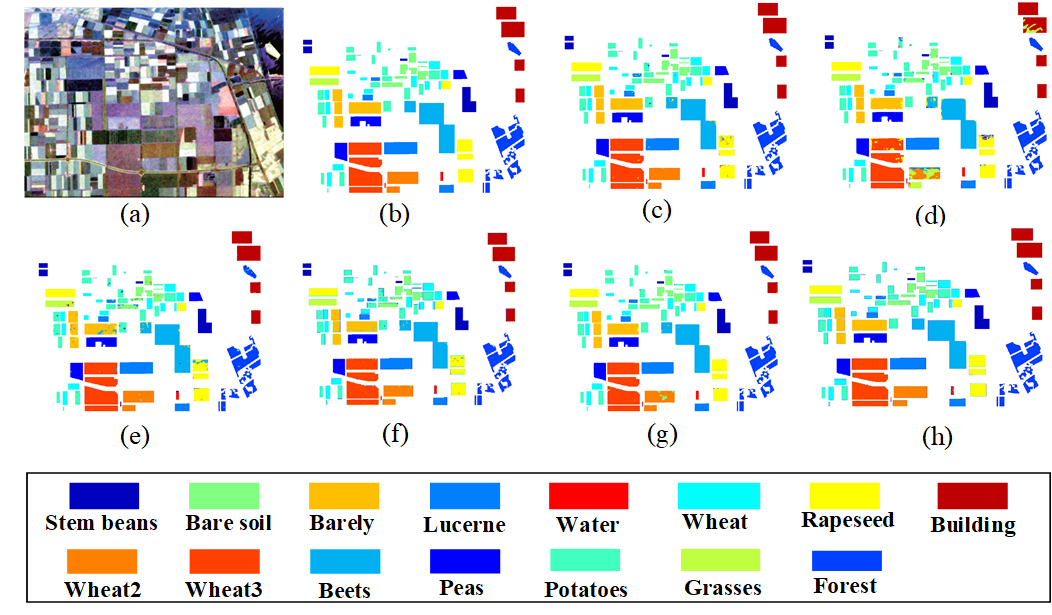}}
	\caption{Classification maps of the Flevoland data set. (a) Ground truth; (b) Super\_RF; (c) CV-CNN; (d) PolMPCNN; (e) DFGCN; (f)SpectralDiff; (g)HybridCVNet ; (h) proposed.}
	\label{fig5}
\end{figure*}

\subsubsection{Flevoland data set}
As shown by the quantitative results in the Table~\ref{t5}, the classification accuracy of different methods varies significantly across land cover types. The Super\_RF method exhibits generally poor performance in most classes, with an overall accuracy (OA) of only 78.70\%, and all evaluation metrics are significantly lower than those of other methods. The CV-CNN method achieves high accuracy in multiple categories such as \textit{water}, \textit{buildings}, and \textit{grasses}, with an OA of 98.96\%, demonstrating strong feature extraction capabilities. The PolMPCNN method performs well in certain classes such as \textit{stem beans} and \textit{forest}, but its OA and Kappa coefficient are only 93.82\% and 93.26\%, respectively, indicating room for improvement. Compared to traditional methods, DFGCN shows a marked improvement with an OA of 97.54\%, and performs particularly well in categories such as \textit{lucerne} and \textit{barely}. SpectralDiff achieves 100\% accuracy in several classes, including \textit{Peas}, \textit{bare soil}, and \textit{wheat2}, and obtains OA of 99.73\%, but its performance in the \textit{buildings} categories is relatively poor. HybridCVNet maintains balanced performance across all metrics and achieves high accuracy in most categories, with OA of 98.98\% and Kappa coefficient of 98.88\%. The proposed MMEFnet achieves the best results across all evaluation metrics, with OA of 99.75\%, AA of 99.40\%, and Kappa coefficient of 99.73\%, demonstrating remarkable advantages in distinguishing various complex land cover types and validating its effectiveness and superiority in PolSAR image classification tasks.

Classification results on the Flevoland dataset generated by the comparison methods and the proposed method are shown in Figs.\ref{fig5}(b)-(h). As illustrated in Fig.\ref{fig5}(b), the Super\_RF method exhibits significant misclassifications across multiple categories, indicating its limited ability to discriminate complex land cover boundaries. The CV-CNN method in Fig.\ref{fig5}(c), leveraging its convolutional structure, reduces some misclassifications to some extent; however, substantial errors remain in the Rapeseed region. The PolMPCNN classification map shown in Fig.\ref{fig5}(d) reveals a notable confusion between the \textit{grass} and \textit{barley} classes. The DFGCN method in Fig.\ref{fig5}(e) integrates global contextual information through graph convolution, effectively suppressing speckle noise, but boundary regions between \textit{rapeseed }and \textit{wheat} still exhibit some ambiguity. SpectralDiff in Fig.\ref{fig5}(f) shows acceptable overall performance, yet a noticeable misclassification occurs in the \textit{buildings} region. In Fig.\ref{fig5}(g), HybridCVNet produces considerable noise in certain farmland areas, which affects the overall visual quality. In contrast, the MMEFnet shown in Fig.\ref{fig5}(h), effectively combining GCN, Riemannian manifold metric, and evidence theory, achieves clearer differentiation between land cover types and achieves the best overall classification performance.

\begin{figure*}[h]
	\centerline{\includegraphics[scale=0.4]{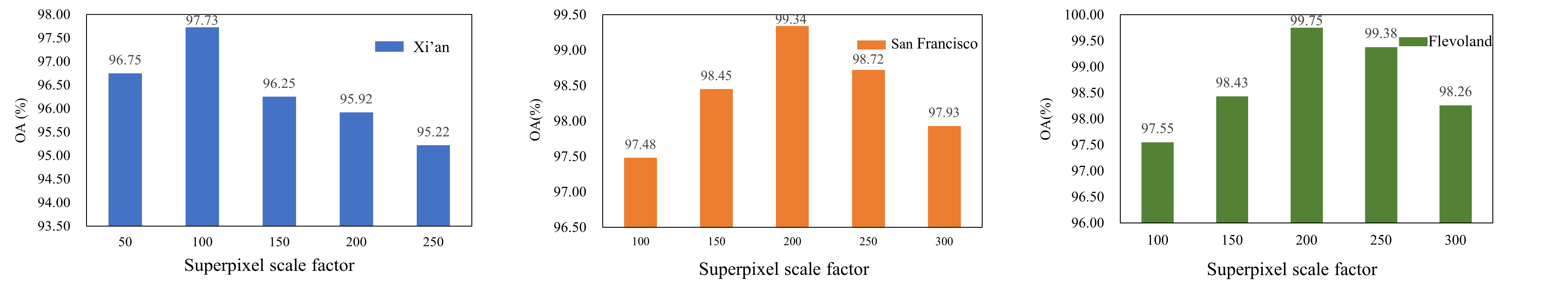}}
	\caption{The effect of superpixel scale parameter on classification accuracy. (a) Xi’an dataset. (b) San Francisco  dataset. (c) Flevoland dataset.}
	\label{fig6}
\end{figure*}

\begin{figure}
	\centerline{\includegraphics[scale=0.2]{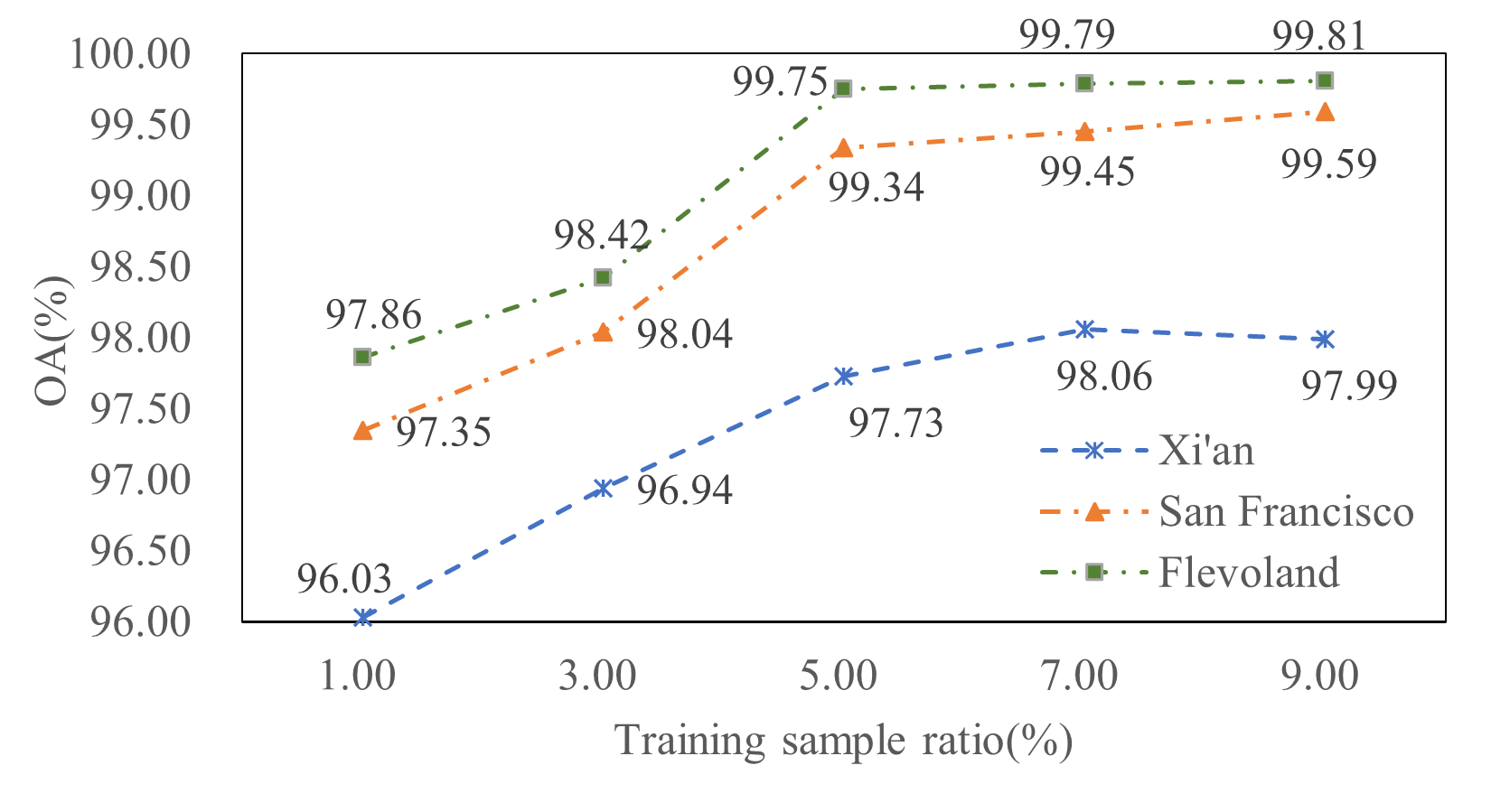}}
	\caption{The effect of superpixel scale parameter on classification accuracy. (a) Xi’an dataset. (b) San Francisco  dataset. (c) Flevoland dataset.}
	\label{fig7}
\end{figure}

\subsection{Ablation Analysis}

Tables \ref{st1}, \ref{st2} and \ref{st3}, respectively, present the ablation results on the Xi’an, San Francisco, and Flevoland datasets. The MMEFnet consists primarily of two parts: Manifold-aware Superpixel-based Graph Convolution Network (sGCN-ME), which is responsible for modeling the graph structure and manifold features within each view, and the Trusted Multiview Evidential Fusion (MEF) module, which performs evidence-level fusion across views. Note that sGCN-ME is essentially an enhanced version of the standard GCN architecture, augmented with manifold-aware metrics to better capture structural and geometric relationships among superpixels.

We first conduct experiments to evaluate the performance of sGCN-ME for each individual view and assess the effectiveness of the manifold-aware metric by comparing it against the baseline sGCN. The models are denoted as "sGCN-v1", "sGCN-v2", sGCN-ME-v1", and "sGCN-ME-v2", corresponding to each view with and without the manifold metric. The impact of the manifold-aware metric can be assessed by comparing "sGCN" with "sGCN-ME". To further evaluate the effectiveness of the multiview fusion strategy, we employ the sGCN-ME model integrated with the TMC module, referred to as "MMEFnet". This model jointly models and fuses the two views at the evidence level. The TMC module facilitates inter-view consistency learning, enabling collaborative optimization and enhanced discriminability of multiview features.

Through these four variants evaluated in the same settings, we systematically quantify individual and joint contributions of manifold enhancement, evidence fusion, and mutual information regularization to the performance of PolSAR multiview classification. We adopt five widely used evaluation metrics to comprehensively assess model performance from multiple perspectives: Overall Accuracy (OA), Average Accuracy (AA), Kappa coefficient, Mean Intersection over Union (MIoU), and weighted F1-Score.
\begin{table}[ht]
\centering
\caption{ABLATION EXPERIMENT RESULTS ON Xi'an Dataset ($\%$)}
\setlength{\tabcolsep}{1pt} 
\begin{tabular}{cccccc}
\hline
\textbf{Method} & \textbf{OA} & \textbf{AA} & \textbf{Kappa} & \textbf{FI-score} & \textbf{MIoU} \\
\hline
sGCN-v1 & 93.59 & 92.10 & 95.01 & 93.12& 91.25 \\
sGCN-ME-v1 & 95.46 & 94.98 & 96.46 & 95.19 & 93.16 \\
sGCN-v2 & 95.83 & 93.45 & 95.25 & 94.35 &95.16 \\
sGCN-ME-v2 &96.41  &95.21  &96.31 &95.56  &94.23 \\
MMEFnet &97.73  &97.08  &96.21  &97.0703  &94.26 \\
\hline
\end{tabular}
\label{st1}
\end{table}
\begin{table}[ht]
\centering
\caption{ABLATION EXPERIMENT RESULTS ON San Francisco Dataset ($\%$)}
\setlength{\tabcolsep}{1pt} 
\begin{tabular}{cccccc}
\hline
\textbf{Method} & \textbf{OA} & \textbf{AA} & \textbf{Kappa} & \textbf{FI-score} & \textbf{MIoU} \\
\hline
sGCN-v1 & 93.49 & 68.99 & 89.56 & 68.99 & 65.50 \\
sGCN-ME-v1 & 97.51 & 95.81 & 96.08& 96.70 & 98.59 \\
sGCN-v2 & 94.73 & 92.59 & 97.65 & 89.32 &82.35 \\
sGCN-ME-v2 &97.54  &94.02  &95.11  &94.23  &92.36 \\
MMEFnet &99.34  &98.44  & 98.27 &98.70  &98.28 \\
\hline
\end{tabular}
\label{st2}
\end{table}

\begin{table}[ht]
\centering
\caption{ABLATION EXPERIMENT RESULTS ON Flevoland  Dataset ($\%$)}
\setlength{\tabcolsep}{1pt} 
\begin{tabular}{cccccc}
\hline
\textbf{Method} & \textbf{OA} & \textbf{AA} & \textbf{Kappa} & \textbf{FI-score} & \textbf{MIoU} \\
\hline
sGCN-v1 & 93.34 & 86.60 & 92.72& 86.60 & 80.35 \\
sGCN-ME-v1 & 95.48 & 92.84 &95.06& 92.28 & 88.06 \\
sGCN-v2 & 96.35 & 89.38 & 96.02& 89.39 &85.31 \\
sGCN-ME-v2 &97.41  & 94.42 &97.17  &94.43  &90.63 \\
MMEFnet &99.75  &99.40  &99.73 &98.37&99.79 \\
\hline
\end{tabular}
\label{st3}
\end{table}

\begin{table*}[htbp]
	\caption{\label{t6} TRAINING AND TESTING TIME OF DIFFERENT METHODS ON XI’AN DATA SET (s)}
	\begin{center}
    \setlength{\tabcolsep}{1.5mm}{
		\begin{tabular}{c c c c c c c c}
			\hline
			Method & Super\_RF & CV-CNN & PolMPCNN & DFGCN & SpectralDiff & HybridCVNet & Proposed \\
			\hline
			Training Time (s) & 60.33 & 3593.12 & 21201.26 & 165.06 & 731.89 & 99.84 & 246.12 \\
			Test Time (s)     & 1.96  & 38.77   & 329.46   & 8.26   & 5.16   & 9.19   & 3.45 \\
			\hline
		\end{tabular}}
	\end{center}
\end{table*}

\subsubsection*{Effect of the Superpixel Scale Parameter $\delta$}

In superpixel segmentation, the scale parameter $\delta$ controls the size of each superpixel, thus affecting both classification performance and computational efficiency. To evaluate the impact of $\delta$ on classification accuracy, we conducted experiments on three datasets and presented the overall accuracy (OA) under different values of $\delta$ using bar charts. For the Xi’an dataset, the best performance was achieved when $\delta = 100$. In contrast, for the larger-scale San Francisco and Flevoland datasets, the optimal accuracy was obtained at $\delta = 200$. Overall, the OA values remain relatively stable within a reasonable range of $\delta$, indicating that the proposed model exhibits a certain degree of robustness to scale variations. The results are illustrated in Figs.~\ref{fig6}(a)–(c).

\subsubsection*{Effect of the Training Sample Ratio}

Comparative experiments with different training sample ratios indicate that when the training sample ratio reaches 5\%, the classification performance on all three datasets has already achieved a high level. Although further increasing the ratio to 7\% and 9\% still brings slight performance improvements, the gains are relatively marginal, exhibiting a clear diminishing trend, as illustrated in Fig.\ref{fig7}. Meanwhile, as the number of training samples increases, the training time grows significantly and computational costs rise. In some cases, the overall accuracy (OA) may even decline slightly. Therefore, considering both performance and efficiency, a training sample ratio of 5\% achieves a satisfactory trade-off and can be regarded as a more desirable choice.

\subsubsection*{Running Time Analysis}
As shown in Table~\ref{t6}, there are significant differences in training and testing time among various methods on the Xi’an dataset. Traditional methods, such as Super\_RF, offer fast training speed but suffer from limited accuracy. In contrast, PolMPCNN demonstrates strong representation capabilities, but its training and inference costs are extremely high, making it less suitable for large-scale applications. In comparison, the proposed MMEFnet achieves a favorable balance between classification accuracy and time efficiency. Its training time is considerably shorter than that of deep models such as PolMPCNN and CV-CNN, taking only 246.12 seconds. The testing time is 3.45 seconds, which is among the best in all deep learning methods. This advantage comes from its efficient feature fusion structure and lightweight inference design. Therefore, MMEFnet demonstrates a well-balanced trade-off between performance and efficiency, making it more practical and promising for real-world applications.


\section{Conclusion}

In this paper, we have proposed a trusted multiview Superpixel-based Graph Convolutional Network (MMEFnet) for PolSAR image classification, which effectively integrates the covariance matrix and multi-feature representations through manifold-aware graph learning and evidence fusion. By embedding each view in its respective manifold space—HPD for the covariance matrix and Grassmann for the multi-feature matrix—and constructing graph structures using kernel-based similarity metrics, the model captures both local geometric and global semantic information. Manifold-aware graph models ensure the learning of geometric separative features from complicated high-dimension data. Furthermore, the introduction of Dempster–Shafer-based evidence fusion enables robust decision-making under uncertainty by resolving conflicts between views and quantifying belief and uncertainty, which effectively reduces misclassification caused by speckle noise. Extensive experiments on multiple real-world PolSAR datasets demonstrate the superiority of the proposed method over existing baselines in terms of precision, robustness, and interpretability. Future work will focus on extending this framework to handle more than two views and enhancing its robustness in more complicated multimodal remote sensing data.

\section*{Acknowledgments}

This work was supported in part by the National Natural Science Foundation of China under Grant 62471387,62272383,62372369, in part by the Youth Innovation Team Research Program Project of Education Department in Shaanxi Province under Grant 23JP111.

\bibliographystyle{IEEEtran}
\bibliography{mybibfile}

\end{document}